\documentclass[letterpaper, 10 pt, conference]{ieeeconf}  

\usepackage{color}
\usepackage{amsfonts,amssymb,amsmath}
\usepackage{graphics}
\usepackage{graphicx}
	 \DeclareGraphicsExtensions{.eps,.jpg,.png}
\usepackage{subfigure}
\usepackage{cite}
\usepackage{todonotes}

\IEEEoverridecommandlockouts                              

\title{\LARGE \bf
Ensemble of Deep Convolutional Neural Networks for Learning to Detect Retinal Vessels in Fundus Images
}

\author{Debapriya Maji$^{*1}$, Anirban Santara$^{*2}$, Pabitra Mitra$^{2}$, Debdoot Sheet$^{2}$
\thanks{$^{1}$~D. Maji is with Texas Instruments Inc.}%
\thanks{$^{2}$~A. Santara, P. Mitra and D. Sheet are with the Indian Institute of Technology Kharagpur, Kharagpur, WB 721302, India.
		{\tt\small anirban$\_$santara@iitkgp.ac.in}}%
\thanks{$^*$ equal contribution}%
}

\begin{document}

\maketitle
\thispagestyle{empty}
\pagestyle{empty}

\begin{abstract}

Vision impairment due to pathological damage of the retina can largely be prevented through periodic screening using fundus color imaging. However the challenge with large scale screening is the inability to exhaustively detect fine blood vessels crucial to disease diagnosis. In this work we present a computational imaging framework using deep  and ensemble learning for reliable detection of blood vessels in fundus color images. An ensemble of deep convolutional neural networks is trained to segment vessel and non-vessel areas of a color fundus image. During inference, the responses of the individual ConvNets of the ensemble are averaged to form the final segmentation. In experimental evaluation with the DRIVE database, we achieve the objective of vessel detection with maximum average accuracy of 94.7\% and area under ROC curve of 0.9283.

\end{abstract}

\begin{keywords}
Computational imaging, deep learning, convolutional neural network, ensemble learning, vessel detection.
\end{keywords}

\section{Introduction}
\label{sec:intro}

Pathological conditions of the retina examined through regular screening~\cite{tuu:1990,ste:2000} can heavily assist prevention of visual blindness. Fundus imaging is the most widely used modality for early screening and detection of such blindness causing diseases like diabetic retinopathy, glucoma, age-related macular degeneration~\cite{hen:2002}, hypertension and stroke induced changes~\cite{won:2001}. Imaging of fundus has largely improved with progress from the film based photography camera to use of electronic imaging sensors; as well as red free imaging, stereo photography, hyperspectral imaging, angiography, etc.~\cite{abr:2010}, thereby reducing inter- and intra-observer reporting variability. Retinal image analysis has also significantly contributed to this technological development~\cite{abr:2010, pat:2006}. Since fundus imaging is predominantly used for first level of abnormality screening, research focus includes: (i) detection and segmentation of retinal structures (vessels, fovea, optic disc), (ii) segmentation of abnormalities, and (iii) quality quantification of images acquired to assess reporting fitness~\cite{abr:2010}.  

\begin{figure}[!t]
      \centering
      \subfigure[Fundus image]{\label{fig:example:16_test}{\includegraphics[height=0.16\textwidth]{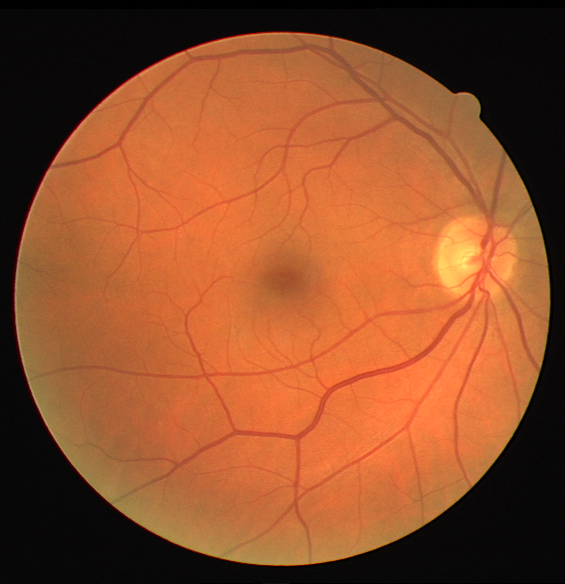}}}
      \subfigure[Ground truth]{\label{fig:example:16_manual1}{\includegraphics[height=0.16\textwidth]{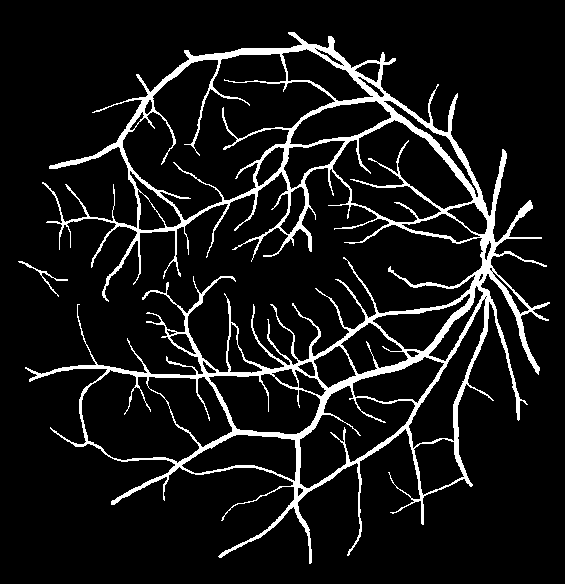}}}
      \subfigure[Detected vessels]{\label{fig:example:16_pred}{\includegraphics[height=0.16\textwidth]{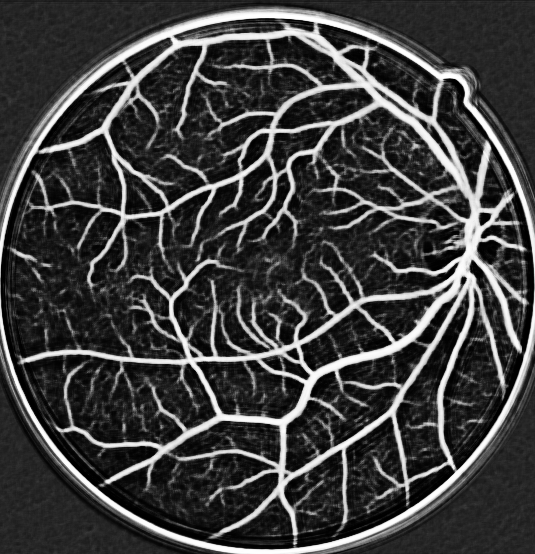}}}
      \subfigure[Proposed Methodology]{\label{fig:plan}{\includegraphics[scale = 0.35]{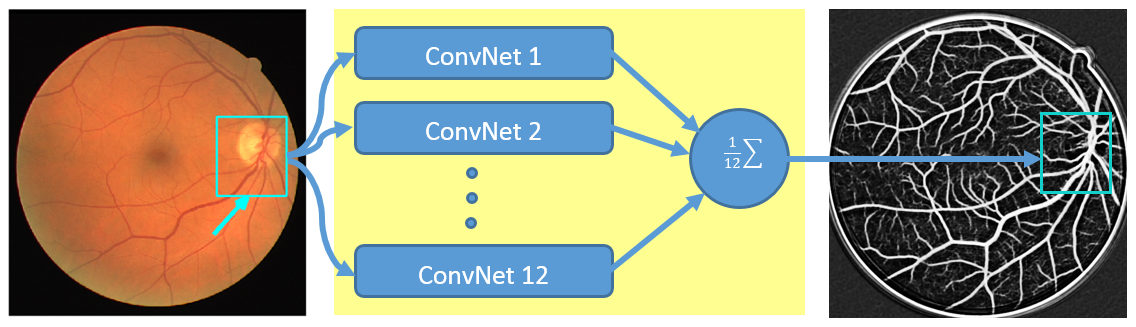}}}
      \caption{Retinal vessel detection using our proposed ensemble of ConvNets. Sample \#16 in~\cite{sta:2004}.}
      \label{fig:example}
\end{figure}

\textbf{Related Work:} The process of clinical reporting of retinal abnormalities is systematic and lesions are reported with respect to their location from vessels or optic disc. Computer assisted diagnosis systems are accordingly being developed to improve the clinical workflow~\cite{abr:2010}. Some of the developments include, assessment of image quality~\cite{geg:2012}; blood vessel detection~\cite{sta:2004}, branch pattern, diameter and vascular tree analysis~\cite{hen:2002,won:2001,son:2006,jan:2012}; followed by reporting of lesions and their location with respect to the vessels~\cite{won:2001,jan:2012}. An important challenge in this context is robust and exhaustive detection of retinal vessels in color fundus images leveraging the potential of computer assisted diagnosis~\cite{abr:2010,nie:2004} and assist in routine screening~\cite{won:2001,hen:2002}.

\textbf{Challenge:} Methods for vessel detection and segmentation ~\cite{abr:2010,pat:2006,sta:2004,son:2006,jan:2012,sheet:2013} predominantly use image filters, vector geometry, statistical distribution studies, and machine learning of low-level features and photon distribution models for vessel detection. Such methods rely on use of handcrafted features or heuristic assumptions for solving the problem and are not generalized to learn pattern attributes from the data itself, thus making them vulnerable to performance subjectivity on account of the method's inherent weaknesses. Recently fully data-driven ,deep learning based models have been proposed~\cite{maji:2015}. However, they are weaker in performance compared to the state of the art methods that use the conventional paradigm. The primary challenge here is to design an end-to-end framework which learns pattern representation from the data without any domain knowledge based heuristic information to identify both coarse and fine vascular structures and is at least at par if not better than the heuristic-based models.

\textbf{Approach:} This paper makes an attempt to ameliorate the issue of subjectivity induced bias in feature representation  by training an ensemble of $12$ Convolutional Neural Networks (ConvNets) ~\cite{lecun:1998} on raw color fundus images to discriminate vessel pixels from non-vessel ones. Fig.~\ref{fig:example} illustrates an example of exhaustive retinal vessel detection using this approach. Each ConvNet has  three convolutional layers and two fully connected layers and is trained independently on randomly selected patches from the training images. At the time of inference, the vesselness-probabilities independently output by each ConvNet are averaged to form the final vesselness probability of each pixel.  

\S \ref{sec:theory} gives a brief theoretical background of the proposed method. The problem statement is formally defined in \S \ref{sec:problem} and the proposed approach is described in \S \ref{sec:soln}. The results of experimental evaluation on DRIVE dataset  have been presented in \S \ref{sec:expt}. The paper is concluded with a summary of the proposed method and a discussion of possible impact of end-to-end deep learning based solutions for medical image analysis in \S \ref{sec:conc}.

\section{Theoretical Background}
\label{sec:theory}
This section introduces some concepts regarding ConvNets and ensemble learning which form the pillars of the proposed solution.

\textbf{Convolutional Neural Networks:} Convolutional neural networks (CNN or ConvNet) are a special category of artificial neural networks designed for processing data with a grid-like structure ~\cite{lecun:1998, bengio-DLBook:2016}. The ConvNet architecture is based on sparse interactions and parameter sharing and is highly effective for efficient learning of spatial invariances in images ~\cite{lecun:2015, mallat:2016}. There are four kinds of layers in a typical ConvNet architecture: convolutional (\texttt{conv}), pooling (\texttt{pool}), fully-connected (\texttt{affine}) and rectifying linear unit (\texttt{ReLU}). Each convolutional layer transforms one set of feature maps into another set of feature maps by convolution with  a set of filters. Mathematically, if $W^l_i$ and $b^l_i$ denote the weights and the bias of the $i^{th}$ filter of the $l^{th}$ convolutional layer and $H^l_i$ be its activation-map, then:
\begin{equation}
H^l_i = H^{l-1} \otimes W^l_i + b^l_i
\end{equation}
where $\otimes$ is the convolution operator.
Pooling layers perform a spatial downsampling of the input feature maps. Pooling helps to make the representation become invariant to small translations of the input. Fully-connected layers are similar to the layers in a vanilla neural network. Let $W^l$ denote the incoming weight matrix and $b^l$, the bias vector of a fully-connected layer, $l$. Then:
\begin{equation}
H^l = flatten(H^{l-1}) * W^l \oplus b^l
\end{equation}
where $flatten(.)$ operator tiles the feature-maps of the input volume along the height, $'*'$ is matrix multiplication and $'\oplus'$ is element-wise addition.
ReLU layers perform a pointwise rectification of the input and correspond to the activation function. For the $i^{th}$ unit of layer $l$:
\begin{equation}
H^l_i = max(H^{l-1}_i,0)
\end{equation}  
In a deep ConvNet, units in the deeper layers indirectly interact with a larger area of the input, thus forming a high level abstraction of the input data. 

\textbf{Ensemble learning:} Ensemble learning is a technique of using multiple models or experts for solving a particular artificial intelligence problem ~\cite{dietterich:2001}. Ensemble methods seek to promote diversity among the models they combine and reduce the problem related to overfitting of the training data. The outputs of the individual models of the ensemble are combined (e.g. by averaging) to form the final prediction. Concretely, if $\{m_1,\dots, m_k\}$ be $k$ models of an ensemble and $p\left(x=y_i|m_j\right)$ is the probability that the input $x$ is classified as $y_i$ under the model $m_j$, then the ensemble predicts:
\begin{equation}
p\left(x=y_i|m_1,\dots,m_k \right)= \frac{1}{k}\sum^k_{j=1}p\left(x=y_i|m_j\right)
\end{equation}
Ensemble learning promotes better generalization and often provides higher accuracy of prediction than the individual models.

\section{Problem Statement}
\label{sec:problem}

Let $\mathcal{I}$ be  an image acquired by the RGB sensor of a color fundus camera. The intensity observed at location $\mathbf{x}$ is denoted by $g(\mathbf{x})$. Let $N(\mathbf{x})$ be a set of pixels in the local neighborhood of $\mathbf{x}$. Let  $\omega=\{\textrm{vessel, non-vascular}\}$ be the set of class labels for the pixel at location $\mathbf{x}$. In a machine learning framework, the probability of finding a tissue of type $\omega$ at location $\mathbf{x}$, $p\left(\omega|\mathcal{I},\mathbf{x}\right)$ is modelled by a class of  functions $\mathcal{H}(\omega|\mathcal{I},N(\mathbf{x}),\mathbf{x},\theta_1,\theta_2)$ where $\theta_1$ is a set of parameters which are learned from the training data and $\theta_2$, a set of hyperparameters tuned using the validation data. In the proposed method, $\mathcal{H}(.)$ is an ensemble of ConvNets, the architecture of which is described next. 

\begin{figure*}[!t]
\centering
	\includegraphics[width=\textwidth]{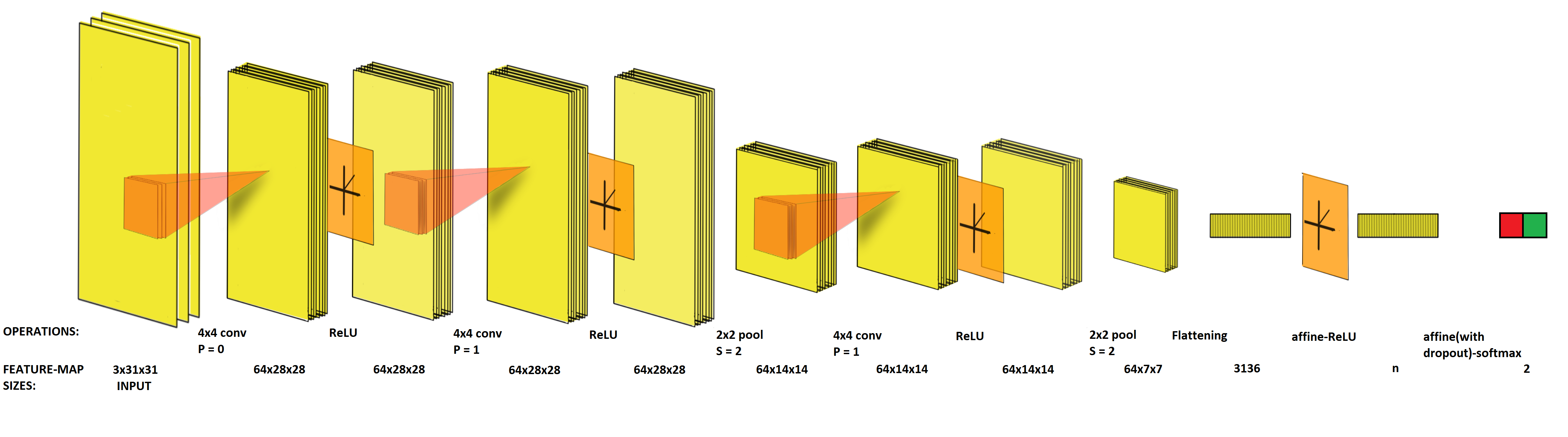}
	\caption{Organization of layers in the ConvNets of the proposed ensemble. S - stride, P - zero padding, n - number of hidden units in penultimate affine layer variable in different ConvNets (see \S \ref{sec:expt} for details)}
	\label{fig:CNN-schematic}
\end{figure*}
\section{Proposed Solution}
\label{sec:soln}
Each layer of a ConvNet transforms one volume of features into another. A volume of features is described as $N\times V\times H$ where $N$ is the number of feature maps of spatial dimension $V\times H$. The input to each ConvNet of the proposed ensemble is a $3\times 31\times 31$ color fundus image patch. The ConvNets have the same organization  of layers which can be described as: \texttt{input- [conv - relu]-[conv - relu - pool] x 2 - affine - relu - [affine with dropout] - softmax}. Fig.\ref{fig:CNN-schematic} gives a schematic diagram of the organization. Each \texttt{conv} layer has receptive field size - $4\times 4$, stride - $1$ and output volume - $64\times 32\times 32$. The \texttt{pool} layers have receptive field size - $2\times 2$ and stride - $2$.  \emph{Dropout}~\cite{srivastava:2014} is a regularization method for neural networks that enforces sparsity, prevents co-adaptation of features and promotes better generalization by forcing a fraction of neurons to be inactive during each episode of learning. The output of the final layer is passed to a \emph{softmax} function which converts the outputs into class probabilities. Let $H^{out}_i$ denote the activation of the $i^{th}$ neuron of the fully-connected output-layer and $P(o_i)$ denote the posterior probability of the $i^{th}$ output class. Then:
\begin{equation}
P(o_i) = softmax(H^{out})_i = \frac{e^{H^{out}_i}}{\sum_j e^{H^{out}_j}}
\end{equation} 

\section{Experimental Results and Discussion}
\label{sec:expt}
This section presents experimental validation of the proposed technique and its performance comparison with earlier methods~\cite{nie:2004}.

\textbf{Dataset:} The ensemble of ConvNets is is evaluated by learning with the DRIVE training set (image id. 21-40) and testing over the DRIVE test set (image id. 1-20)\footnote{DRIVE dataset: http://www.isi.uu.nl/Research/Databases/DRIVE/}.

\textbf{Learning mechanism:} Each ConvNet is trained independently on a set of $60000$ randomly chosen $3\times31\times31$ patches. Learning rate and annealing rate were kept constant across models at $5e-4$ and $0.95$ respectively. Dropout probability, $L_2$ regularization coefficient and number of hidden units in the penultimate affine layer of the different models were sampled respectively from $\mathcal{U}\left([0.5,0.9]\right)$, $\mathcal{U}\left([1e-3,2.5e-3]\right)$ and $\mathcal{U}\left(\{128, 256, 512\}\right)$ where $\mathcal{U}(.)$ denotes uniform probability distribution over a given range. The models were trained using RMSProp algorithm ~\cite{hinton:notes} with minibatch size $200$.

\begin{figure}[!t]
      \centering
      \subfigure[Image \# 5]{\label{fig:result:5_arrow}{\includegraphics[height=0.12\textwidth]{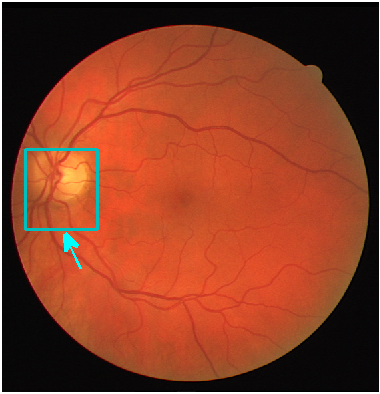}}}
      \subfigure[Mag. view]{\label{fig:result:5_zoomed_coarse}{\includegraphics[height=0.12\textwidth]{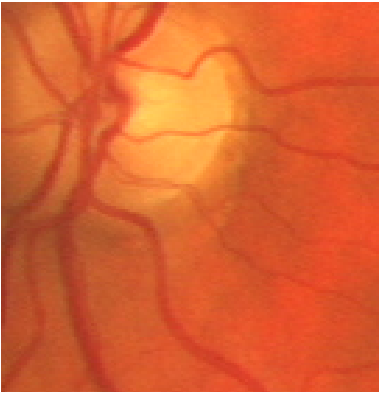}}}
      \subfigure[Ground truth]{\label{fig:result:5_coarse_gt}{\includegraphics[height=0.12\textwidth]{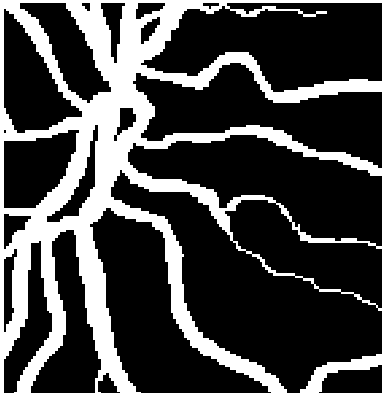}}}
      \subfigure[Detected]{\label{fig:result:5_det_coarse}{\includegraphics[height=0.12\textwidth]{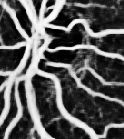}}} 

      \subfigure[Image \# 16]{\label{fig:result:16_coarse_arrow}{\includegraphics[height=0.12\textwidth]{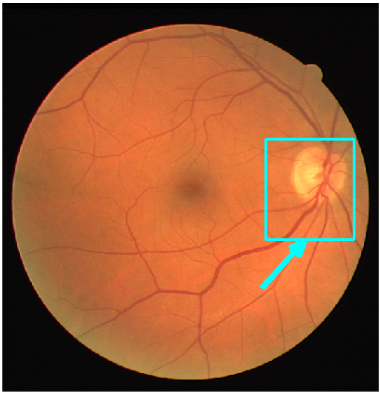}}}
      \subfigure[Mag. view]{\label{fig:result:16_zoomed_coarse}{\includegraphics[height=0.12\textwidth]{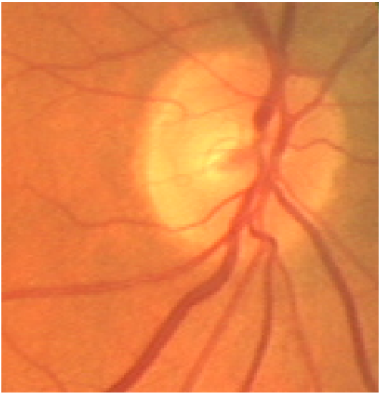}}}
      \subfigure[Ground truth]{\label{fig:result:16_coarse_gtr}{\includegraphics[height=0.12\textwidth]{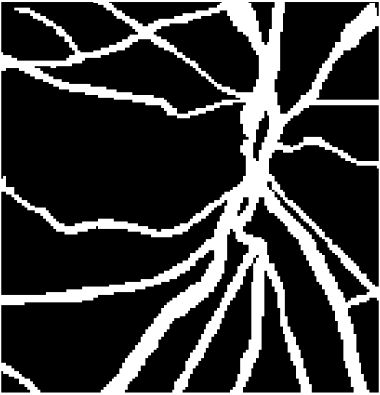}}}
      \subfigure[Detected]{\label{fig:result:16_coarse_det}{\includegraphics[height=0.12\textwidth]{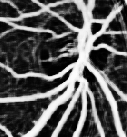}}} 

      \subfigure[Image \# 16]{\label{fig:result:16_fine_arrow}{\includegraphics[height=0.12\textwidth]{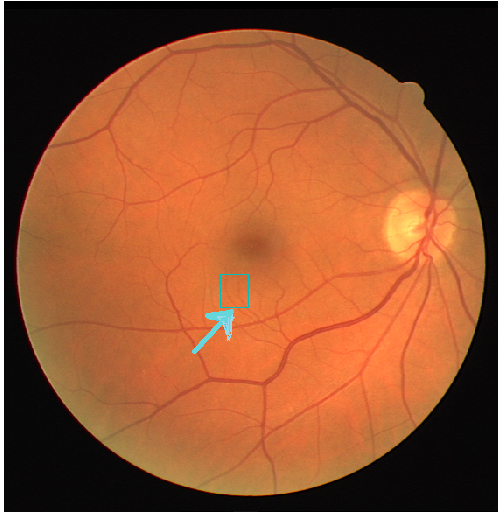}}}
      \subfigure[Mag. view]{\label{fig:result:16_zoomed_fine}{\includegraphics[height=0.12\textwidth]{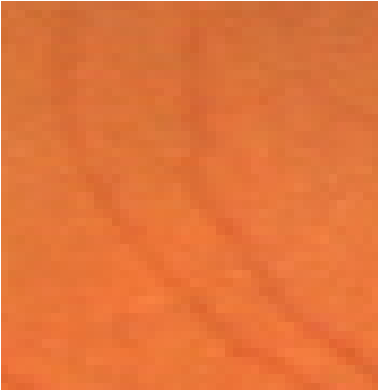}}}
      \subfigure[Ground truth]{\label{fig:result:16_gt_fine}{\includegraphics[height=0.12\textwidth]{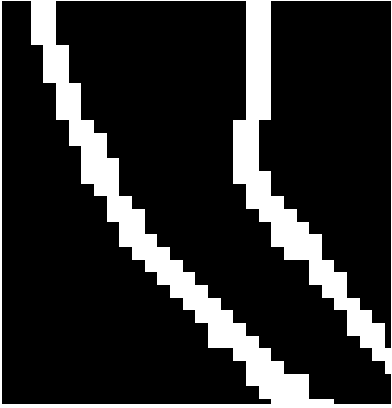}}}
      \subfigure[Detected]{\label{fig:result:16_det_fine}{\includegraphics[height=0.12\textwidth]{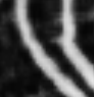}}} 

      \caption{Detection of (a-d) coarse vessels in Sample \#5 and both (e-h) coarse and (i-l) fine vessels in Sample \#16 in~\cite{sta:2004}.}
      \label{fig:results}
\end{figure}

\textbf{Performance assessment:} Table~\ref{tab:perftest} presents the accuracy and consistency of detection in comparison with those reported in earlier techniques. It is clearly evident that although our approach does not have the highest accuracy as compared with other methods, it does exhibit superior performance than the previously proposed deep learning based method for learning vessel representations from data \cite{maji:2015}. The kappa score being a study of observer consistency indicates sensitivity of the technique to detect both coarse and fine vessels as desired. Typical response to detection of both coarse and fine vessels are presented in Fig.~\ref{fig:results}.

\begin{table}[!th]
\centering
\caption{Performance comparison of different algorithms.}
\label{tab:perftest}
\begin{tabular}{|l|c|c|}
\hline
 & Max. avg. Accuracy & Kappa \\
\hline
\textbf{Proposed method} & \textbf{0.9470} & \textbf{0.7031} \\
Maji et al.~\cite{maji:2015} & 0.9327 & 0.6287 \\
Sheet et al.~\cite{sheet:2013} & 0.9766 & 0.8213 \\
Second observer & 0.9473 & 0.7589 \\
Staal et al.~\cite{sta:2004}. & 0.9422 & - \\
Niemeijer et al. & 0.9416 & 0.7145 \\
Zana et al. & 0.9377 & 0.6971 \\
Jiang et al. & 0.9212 & 0.6399 \\
Mart\'{i}nez-P\'{e}rez et al. & 0.9181 & 0.6389 \\
Chaudhuri et al. & 0.8773 & 0.3357 \\
\hline
\end{tabular}
\end{table}

\begin{figure}
\centering
\includegraphics[width=0.5\textwidth]{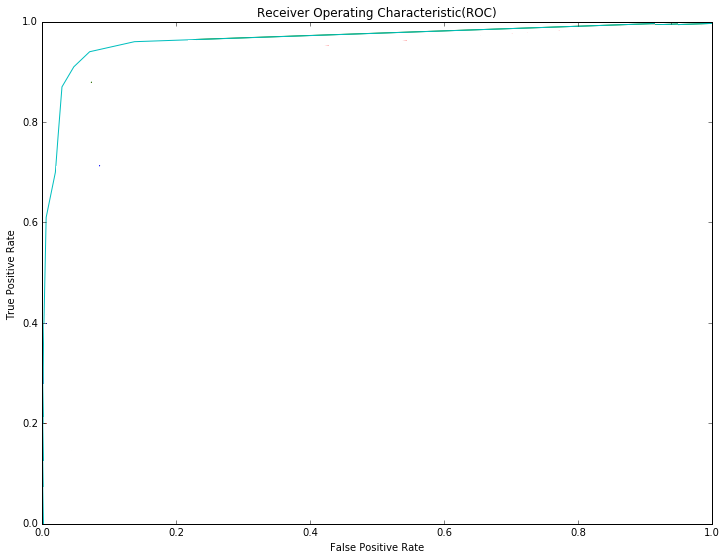}
\caption{Receiver Operating Characteristic (ROC) curve}
\label{fig:ROC}
\end{figure}

Fig: \ref{fig:ROC} gives the receiver operating characteristic (ROC) curve of the proposed method. Area under ROC curve obtained is $0.9283$.

\section{Conclusion}
\label{sec:conc}

This paper presents a ConvNet-ensemble based framework for processing color fundus images for detection of coarse and fine vessels. The method is evaluated experimentally on the DRIVE dataset. The remarkable ability of ConvNets to recognize images and that of ensemble learning at generalization is leveraged to design a heuristics independent, data driven approach to analyzing medical images. This presents a feasible solution to subjectivity induced bias in medical image analysis. This is an improvement of our previous work on data-driven analysis of fundus images \cite{maji:2015}. This approach in general also provides a strong alternative approach to solve complex medical data analysis problem through deep learning combined with the power of ensemble learning.

\bibliographystyle{IEEEbib}
\small
\bibliography{refs}

\end{document}